\documentclass[runningheads,a4paper]{llncs}

\usepackage{amssymb}
\usepackage{graphicx}
\usepackage{hyperref}
\usepackage{url}
\usepackage{amsmath}
\usepackage{tikzsymbols}
\usepackage{tabularx}
\usepackage{caption,subcaption}
\usepackage{fancyhdr}
\fancypagestyle{firstpage}
{
    \fancyhead[L]{In the proceedings of CICLING 2018}    
    \fancyhead[R]{}
}%
\urldef{\mailsa}\path|{nurendra.choudhary, rajat.singh}@research.iiit.ac.in|
\urldef{\mailsc}\path|ishita.bindlish@students.iiit.ac.in|
\urldef{\mailsb}\path|m.shrivastava@iiit.ac.in|

\begin{document}

\mainmatter  
\title{Neural Network Architecture for Credibility Assessment of Textual Claims}

\titlerunning{Neural Network Architecture for Credibility Assessment}

%
%


%
%

\makeatletter
\newcommand{\printfnsymbol}[1]{%
  \textsuperscript{\@fnsymbol{#1}}%
}
\makeatother
\author{Nurendra Choudhary\thanks{These authors have contributed equally to this work.}, Rajat Singh\printfnsymbol{1}, Ishita Bindlish \and Manish Shrivastava}

\institute{Language Technologies Research Centre (LTRC)\\ Kohli Center on Intelligent Systems (KCIS)\\ International Institute of Information Technology, Hyderabad, India
\mailsa\\
\mailsc\\
\mailsb\\
}
\maketitle

\thispagestyle{firstpage}

\begin{abstract}
Text articles with false claims, especially news, have recently become aggravating for the Internet users. These articles are in wide circulation and readers face difficulty discerning fact from fiction. Previous work on credibility assessment has focused on factual analysis and linguistic features. The task's main challenge is the distinction between the features of true and false articles. In this paper, we propose a novel approach called Credibility Outcome (CREDO) which aims at scoring the credibility of an article in an open domain setting.

CREDO consists of different modules for capturing various features responsible for the credibility of an article. These features includes credibility of the article's source and author, semantic similarity between the article and related credible articles retrieved from a knowledge base, and sentiments conveyed by the article. A neural network architecture learns the contribution of each of these modules to the overall credibility of an article. Experiments on Snopes dataset reveals that CREDO outperforms the state-of-the-art approaches based on linguistic features.
\end{abstract}
\section{Introduction}
\textit{``Fake news is a type of hoax or deliberate spread of misinformation, be it via the traditional news media or social media, with the intent to mislead, in order to gain financially or politically."} \cite{leonhardt2017trump}

The fake articles do not base their information on facts but use convenient, seemingly true, logical inferences to make readers trust the news. In recent times, the difference between real and fake news articles has become bleak. Verification of the information needs checking for reliable sources. Most readers find the task cumbersome and hence don't perform it. Therefore, automated systems that detect such articles are a major requirement.

Correction of misinformation is difficult. Fabricated news persists because of people casually inferring based on available information about a given event. As a result, false information continues to influence opinions, beliefs and attitudes even after being debunked, unless replaced by an alternate factual explanation. The situation deteriorates when the fake articles have a financial or political agenda. The manner in which these articles shape public opinion affects the society as a whole, making it a very serious problem.

The modules of CREDO's (Credibility Outcome) include keyword extraction, document retrieval, author credibility scores, website trust scores, semantic similarity and sentiment analysis. Keyword extraction module chooses significant words in the given input article which, in the document retrieval phase, fetches documents from a knowledge base, primarily, news pages and Wikipedia. Querying the entire document gives us negative results and is not efficient, considering the long queries. Subsequent phases learn trust scores of the source websites of the retrieved articles and credibility of their authors. To ensure that our retrieved document and the given article are not dissimilar, we also compute semantic similarity. An article is a sequence of words. Semantic similarity modules need to include a method that captures this sequence of words and also learn a distance metric between them. Long Short-Term Memory (LSTM) based recurrent neural networks have proven helpful in capturing sequences \cite{palangi2016deep}. Also, siamese networks have shown promising results in distance-based learning methods \cite{bromley1994signature}. Hence, we use a combination of these by utilizing a bidirectional LSTM to map articles to a semantic space in conjunction with a siamese network to learn the similarity metric between them.

A factual article has more probability of being neutral \cite{nasukawa2003sentiment}. Hence, a system to capture this feature is essential. To tackle the problem, we use a sentiment analysis \cite{pang2008opinion} tool to evaluate the neutrality of a given article.

The rest of the paper is organized as follows. We discuss previous approaches in the field, motivating us for the task in section \ref{sec:related work}. In section \ref{sec:credo}, we discuss the pipeline and individually explore each module of CREDO. Section \ref{sec:dataset and baselines} describes the datasets and baselines considered for the task. Section \ref{sec:experiments} explains our experiments and their evaluation. Finally, section \ref{sec:conclusion} concludes the paper.

\section{Related Work}
\label{sec:related work}
The work connects a significant number of research areas, including but not limited to automatic fact checking, rumour detection, sentiment analysis, semantic similarity and credibility analysis.

\cite{weikum2017computers} discusses the belief system of computers and credibility analysis' necessity. \cite{vlachos2014fact} discusses fact checking, its definition and motivations, posing it as a classification task. \cite{castillo2011information} proposes credibility analysis in a social media context, thereby using features that describe users' posting behaviour. Joint model based on CRF \cite{popat2017truth} employs linguistic features like assertive verbs, discourse markers among others to establish a common style of writing across fake articles and news. This approach proves effective and achieves more than 80\% accuracy on snopes dataset.
 
CoTruth \cite{liu2017truth} uses a similar approach in employing linguistic features as a joint model based on CRF \cite{popat2017truth} but does not perform a credibility check on the author and website.

The above approaches model the writing style. However, they do not consider the available knowledge bases. The writing style also depends on the content's authors and the website's type. A unified model to capture the style without taking into account the information of authors and website will undoubtedly be fragile.

In semantic similarity, we use siamese networks, to compute the similarity between two sentences. \cite{bromley1994signature} first introduced siamese networks in 1994 to solve the problem of signature verification. Since then, there have been attempts to understand their relevance in the context of sentence similarity. In SCQA model \cite{das2016together}, siamese networks solve the task of community question answering and in DSSM model \cite{huang2013learning}, they handle the task of website ranking. The above methods use the siamese network based on the task. Here, the task is the semantic similarity. Hence, we use LSTM models \cite{palangi2016deep} to project our articles in the semantic space to learn a similarity metric between them.
\begin {table}
\centering
\begin{tabular}{ m{28em} | m{8.5em} } 
Article & Keywords\\\hline
In May 1946, Einstein made a rare public appearance outside of Princeton, New Jersey, when he traveled to the campus of Pennsylvania's Lincoln University, the United States' first degree-granting black university, to take part in a ceremony conferring upon him the honorary degree of doctor of laws. & [ (`degree-granting black university', 8.5), (`lincoln university', 4.5), (`ceremony conferring', 4.0)]  \\\hline
There are two fatal problems with the JATO story. First, anybody who understood the extreme forces involved well enough to attach a JATO unit to a car so that it would keep the car going in a straight line would probably know better than to do it in the first place.& [ (`extreme forces involved', 9.0), (`fatal problems', 4.0), (`straight line', 4.0)] \\
\end{tabular}
\caption{Result of Keyword Extraction}
\label{tab:keyword}
\end{table}

\section{CREDO: Methodology and Architecture} \label{sec:credo}
CREDO or Credibility Outcome is a neural network architecture combining information retrieval, semantic similarity, and sentiment analysis. The following subsections explain the architecture's modules.
\subsection{Keyword extraction and Document retrieval}
\label{sec:keyword extraction and document retrieval}
Keywords provide us with most of the significant information contained in the article. Hence, keyword extraction becomes an essential initial step for document retrieval. The module creates a query using keywords of input text to retrieve relevant documents. We use Rapid Automatic Keyword Extraction (RAKE) described in \cite{rose2010automatic}. RAKE involves three sub-modules -
\begin{enumerate}
\item \textbf{Candidate selection} extracts all potential keywords (words, phrases and concepts).
\item \textbf{Properties calculation} measures the candidate's indication of being a keyword. For example, a likely keyword is a candidate in the title of a book.
\item \textbf{Scoring and selecting keywords} scores the candidates by either combining the properties into a formula or utilizes a machine learning technique to determine the probability of a candidate being a keyword. The final set are the keywords with a score above an empirical threshold. This assigns the keywords to an article, with their respective scores representing their importance in the article. Some examples of the extracted keywords are given in table \ref{tab:keyword}.
\end{enumerate}

Keyword extraction and document retrieval module uses the keywords to query over the knowledge bases (proprietary set of indexed documents, such as Wikipedia, or a Web search engine like Google search and Bing search) using an information-retrieval system. The result of this document retrieval stage is a set of relevant documents.

Although the set of documents is relevance-ranked, the top entities are probably not credible enough. Keyword relevance is not an appropriate ranking metric for a credibility scoring system. The existence of a highly relevant and large document full of fake news is easily possible. Therefore, we rely on other modules to check document credibility.

\subsection{Website and Author Trust Scores}
\label{sec:website and author trust scores}
In this module, given an article's source, the Web of Trust API\footnote{\url{www.mywot.com}} initializes the credibility score of the website. Web of Trust, a website reputation and review service, provides information about the trustworthiness of a website. The information's basis is a combination of crowd-sourced reviews and identification of networks involved in malware distribution.

With the scores from Web of Trust, we perform a logarithmic gradient descent based on the tag of the input in the dataset. i.e.
\begin{equation}
\label{eq:wts}
WTS (w_{i+1})=
\begin{cases}
 (i+\log (1-t (w_{i})))\times WTS (w_{i}),t (w_{i})\leq0.5\\
 (i+\log (1+t (w_{i})))\times WTS (w_{i}),t (w_{i})>0.5
\end{cases}
\end{equation}
where $WTS (w_{i})$ is the web trust score of website $w$ for its $i^{th}$ instance in the sample and $t (w_{i})$ represents the tag of the $i^{th}$ article.

Similarly, initial articles of an author decide his score for the present one. An initial score of 0.5 is set for each author. If the author is anonymous or irretrievable from the dataset, the probability of an author being credible and not credible is assumed to be equal and hence, the ACS score is set to 0.5. The metric for author credibility score is:
\begin{equation}
\label{eq:acs}
ACS (a_{i+1}) =
\begin{cases}
 (i+\log (1-t (a_{i})))\times ACS (a_{i}),t (a_{i})\leq0.5\\
 (i+\log (1+t (a_{i})))\times ACS (a_{i}),t (a_{i})>0.5
\end{cases}
\end{equation}
where $ACS (a_{i})$ is the author credibility score of author $a$ at his $i^{th}$ article in the sample and $t (a_{i})$ denotes the tag of the $i^{th}$ article.

New training instances dynamically update these calculated scores according to their contribution to the final credibility Score of the source website and the article's author.

\subsection{Document Summarization}
\label{sec:document summarization}
Only specific sections of the document demonstrate the significant relevant information in a text article. Hence, based on the keywords and TextRank algorithm \cite{mihalcea2004textrank} with BM25 ranking function \cite{barrios2016variations} (implementation by Gensim\footnote{\url{https://radimrehurek.com/gensim/summarization/summariser.html}}), we summarize the document by extracting the top \textit{k} sentences based on the algorithm's ranking. We choose \textit{k} such that the summary size is  approximately the same as that of the input article.
Summarization of large input articles emphasize the most significant information, while at the same time, simplifying keyword extraction.
\begin{figure}
\centering
\begin{minipage}{.49\textwidth}
  \centering
  \includegraphics[width=.87\linewidth]{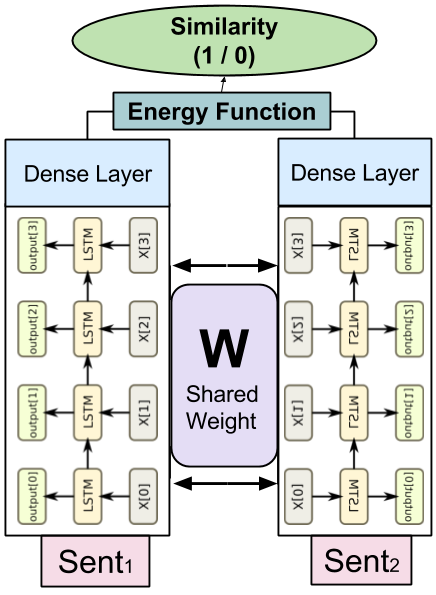}
  \captionof{figure}{Siamese Architecture for Semantic Similarity}
  \label{fig:siamese}
\end{minipage}%
\begin{minipage}{.48\textwidth}
  \centering
  \includegraphics[width=\linewidth]{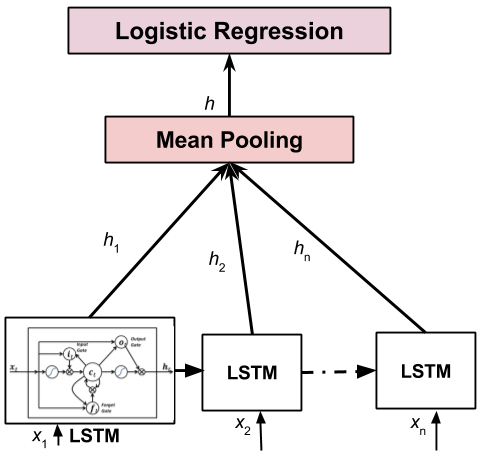}
  \captionof{figure}{LSTM RNN for Sentiment Classification}
  \label{fig:sentiment}
\end{minipage}
\end{figure}

\subsection{Semantic Similarity}
\label{sec:semantic similarity}
Semantic similarity is of paramount importance in the scoring task. Given the summarized versions of the original input article and the documents retrieved in the previous steps, we calculate the semantic similarity score between them. This is necessary because only keywords influence the documents retrieval from the knowledge base. So, it is possible that the keywords match the negative meaning or the document retrieved and the original input are unrelated, apart from just some words.

For example, a web search on the query ``Donald Trump is the president of India" results in titles - ``How US President Donald Trump's world-view affects India", ``US President Donald Trump to speak to PM Modi as India hopes to build on US ties" etc, which are considered good evidences by the system but are semantically different.

In this model, siamese architecture with LSTMs calculates the semantic similarity, similar to how the DSSM model \cite{huang2013learning} computes semantic similarity with primarily two differences:
\begin{itemize}
\item A fully connected neural network is the DSSM model's basis, whereas, we employ an LSTM here instead. LSTMs capture sequential information data types like texts, in the sense that they capture order and history.
\item The basic unit in their model constitutes character n-grams, whereas we use tokens or words. We do not need character based vectors because news articles are not susceptible to spelling errors and out of vocabulary words.
\end{itemize}

As illustrated in figure \ref{fig:siamese}, the architecture consists of twin LSTM networks with a similarity-based energy function on the top. The LSTM networks map the article to a vector in the semantic space and the energy function learns the similarity between them. The similarity metric is learned using contrastive learning. The sentences with similar meaning are labeled +1 (positive samples) and dissimilar meaning sentences are labeled -1 (negative samples). 
The results of semantic similarity are explained later in Section \ref{sec:dependencyonmodules}.
\subsection{Sentiment Analysis}
\label{sec:sentiment analysis}
Sensationalized or opinionated news articles are sentiment heavy whereas factual articles are more objective and neutral. To establish a relation between credibility and sentiment of an article, we employ a sentiment analysis tool based on LSTM (Long Short-Term Memory) networks. LSTM networks show remarkable results in learning sequential information like texts \cite{sundermeyer2012lstm}; \cite{greff2017lstm}. As depicted in Figure \ref{fig:sentiment}, we use an LSTM model with single layer. It transforms the input sequence $<x_{i}>$ to a representation sequence $<h_{i}>$. Then a pooling layer averages representations $h_{i}$s over all time steps, resulting in a single representation. Finally, we train a logistic regression model on the representations whose target is the sentiment label (+1 or -1) corresponding to the input sequence. 
Our motivation to use an LSTM based sentiment classification model without using any manual linguistic features is to avoid any linguistic encoding throughout CREDO and rely essentially on machine intelligence. Section \ref{sec:dependencyonmodules} describes the influence of this module over the system.
\subsection{Ensemble of the Modules}
\label{sec:ensemble of the modules}
\begin{figure} 
\centering
\includegraphics[width=.8\textwidth]{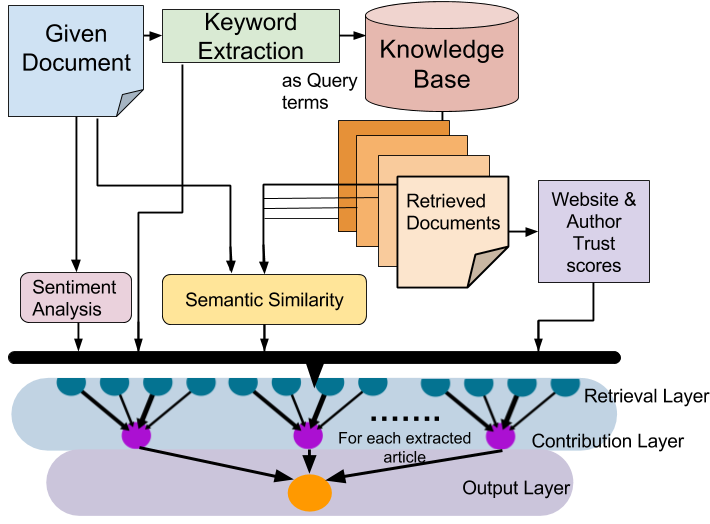}
\centering
\caption{Architecture of CREDO}
\label{fig:neuralnetwork}
\end{figure}
Taking Naive Bayes assumption (or conditional independence assumption), we consider all the above features independent of each other. They together contribute to the article's overall credibility. So, credibility contribution of each article is given by:
\begin{multline}
\label{credibility contribution}
CC (a_{r},a_{i}) = (w_{1}\times (\sum_{i \in k} kscore (i))) + (w_{2}\times{ACS} + w_{3}\times{WTS})\\
+ (w_{4}\times{NS}) + ((w_{5})\times{SS (a_{i},a_{r})})\\
\text{given that, $w_{1}+w_{2}+w_{3}+w_{4}+w_{5} = 1$ \& $w_{1},w_{2},w_{3},w_{4},w_{5} > 0$}
\end{multline}
where $a_{r}$, $a_{i}$ are the retrieved article and given input article respectively, $CC (a_{r},a_{i})$ represents the credibility contribution of an article $a_{r}$ with respect to the given article $a_{i}$, $k$ represents the keywords in the article $a_{r}$ $kscore (i)$ represents the keyword scores of the $i^{th}$ keyword, $ACS$ represents the author's credibility score derived from equation \ref{eq:acs}, $WTS$ represents the trust score of the website derived from equation \ref{eq:wts}, $SS (a_{i},a_{r})$ represents the similarity score between the input and retrieved article and $NS$ represents neutral sentiment's value of $a_{i}$, scaled in 0-1. $w_{1}, w_{2}, w_{3}, w_{4}, w_{5}$ are the respective weight parameters given to these measures in the overall credibility.

Since we retrieve multiple articles, we repeat the same process for every one. We assume that the credibility of all the retrieved articles, weighted with a function of their respective retrieval ranks, will contribute to the final credibility score for the given input article.
\begin{equation}\label{credibility}
Credo (a_{i}) = \frac{\sum_{r\in R} (e^{1-\frac{rank (a_{r}}{n})}\times CC (a_{r}))}{\sum_{r\in R}e^{1-\frac{rank (a_{r})}{n}}}\\
\end{equation}
where $Credo (a_{i})$ is the input article's overall credibility $a_{i}$, $R$ denotes the set of retrieved documents, $n$ denotes the number of documents retrieved, $rank (a_{r})$ represents the retrieval rank of article $a_{r}$ and $CC (a_{r})$ denotes the credibility contribution of $a_{r}$ given by equation \ref{credibility contribution}. Equation \ref{credibility} gives us the input article's credibility in the range $[0,1]$. Figure \ref{fig:neuralnetwork} details the overall architecture of CREDO.

The weights in the equations \ref{credibility contribution} and \ref{credibility} define the parameters for the module's linear combination scores. Various types of classifiers learn these parameters. Section \ref{sec:experiments} explains the different trials.
The weights learned assigns the credibility score to any new input text article.

\section{Dataset \& Baselines}
\label{sec:dataset and baselines}
This section explains the choice of Dataset for the task and also describes the construction of baselines from the previous approaches to compare with CREDO.
\begin{table*}[t]\centering
  \begin{tabular}{c|c|c|c|c|c|c|c|c}
   & & True & False & Macro- & & Fake & Fake & Fake \\
   & Overall  & Claims & Claims & averaged & & Claims & Claims & Claims \\ 
Experiment & Accuracy & Accuracy & Accuracy & Accuracy & AUC & Precision & Recall & F1-score \\     \hline
\textbf{CRF} & 81.39 & 83.21 & 80.78 & 82.00 & 0.88 & 0.93 & 0.81 & 0.87\\
\textbf{LG + SR} & 71.96 & 75.43 & 70.77 & 73.10 & 0.80 & 0.89 & 0.71 & 0.79\\
\hline
\textbf{RBF-SVM} & 82.8 & 85.7 & \textbf{81.5} & 83.6 & \textbf{0.87} & \textbf{0.94} & \textbf{0.85} & \textbf{0.89}\\
\textbf{MLP-NN} & \textbf{83.3} & \textbf{86.2} & 81.2 & \textbf{83.7} & 0.83 & 0.92 & 0.84 & 0.88\\\hline
\textbf{Multi-Class} & 56.4 & 59.8 & 55.2 & 57.5 & 0.63 & 0.63 & 0.53 & 0.57\\
  \end{tabular}
  \caption{Comparison of CREDO with baselines (LG+SR and CRF).}
  \label{tab:table1} 
\end{table*}
\subsection{Dataset}\label{sec:dataset}
The following datasets are considered for training and evaluation of CREDO.

\textbf{Snopes Dataset:}
The dataset used for comparative analysis of the Credibility Scoring Algorithm is Snopes Dataset. Previous approaches to the problem \cite{popat2017truth}, \cite{popat2016credibility} use the same dataset.

Each article on Snopes verifies a single claim. The Snopes' editors assign a manual credibility verdict to each such claim: True or False. Few of the claims have labels \textit{Mostly True}, \textit{Mostly False}, \textit{Partially True} or \textit{Partially False}. A description of the editors' arrival to the claim accompanies the credibility verdict (e.g., collected from a Facebook post or received by email etc.) - an \textit{Origins} section describes the claim's origin, and a \textit{Description} section justifies the verdict. We consider \textit{True} and \textit{Mostly True} as \textit{True} claims. Similarly, we consider \textit{False} and \textit{Mostly False} as \textit{False}. The dataset has 4856 total claims with 1277 \textit{True} claims and 3579 \textit{False} claims.

\textbf{SemEval 2016 Dataset:}
For evaluating the Semantic Similarity module, we adopt the standard SemEval-2016 Task1 - English Semantic Textual Similarity (STS) Dataset \footnote{\url{http://alt.qcri.org/semeval2016/task1/index.php?id=data-and-tools}}. The dataset consists of sentence pairs with their similarity scores in binary.

The dataset has five different sentences' types drawn from various sources. The dataset of \textit{Answer-Answer} and \textit{Question-Question} is taken from Stack Exchange Q\&A Forums\footnote{\url{https://archive.org/details/stackexchange}}, \textit{Headlines} data is taken from Europe Media Monitor (EMM) \cite{best2005europe}, data of \textit{Plagiarism} is taken from corpus of plagiarized short answers \cite{clough2011developing} and data of \textit{Postediting} is taken from WMT quality estimation shared task \cite{Callison-Burch:2012:FWS:2393015.2393018}.

\subsection{Baselines}
\label{sec:baselines}
We test CREDO system against the previous approaches in the problem \cite{popat2017truth} and \cite{popat2016credibility} and evaluate it against the same metrics.
\begin{itemize}
\item \textbf{LG + SR:} This approach uses language stylistic features and source reliability to determine the credibility tag of an article using a distant supervision model. \cite{popat2016credibility}
\item \textbf{CRF}: This approach improves upon the \textit{LG+SR} and attempts to solve the problem using Conditional Random Fields (CRF) dependent on web-sources, articles, claims and claim credibility labels. \cite{popat2017truth}
\end{itemize}

\section{Experiments}\label{sec:experiments}
We conducted an array of experiments to understand the effect of different modules on the system's overall performance. For this, we first excluded the modules and then observed the effect of this exclusion on the system's evaluation metrics. We also conducted experiments to understand the effect of different classifiers on the system. We calculated the evaluation metrics for different classifiers and attempted to understand the reasons for the difference in results. Semantic similarity, being a major module, was evaluated independently to improve results.
\begin{table*}[t]\centering
  \begin{tabular}{m{4.8em}|cccccccc}
   & & True & False & Macro- & & Fake & Fake & Fake \\
Excluding & Overall  & Claims & Claims & averaged & & Claims & Claims & Claims \\ 
Modules & Accuracy & Accuracy & Accuracy & Accuracy & AUC & Precision & Recall & F1-score \\\hline
CREDO \textit{w/o} ACS & 80.5 & 82.3 & 81.3 & 81.8 & 0.83 & 0.90 & 0.82 & 0.86\\
CREDO \textit{w/o} WTS & 81.8 & 84.7 & 80.4 & 82.55 & 0.85 & 0.92 & 0.83 & 0.87 \\
CREDO \textit{w/o} SS & 50.6 & 53.4 & 50.4 & 51.9 & 0.53 & 0.58 & 0.56 & 0.57 \\
CREDO \textit{w/o} SA & 76.2 & 78.4 & 75.4 & 76.9 & 0.76 & 0.77 & 0.76 & 0.76 \\
CREDO & \textbf{83.3} & \textbf{86.2} & \textbf{81.2} & \textbf{83.7} & \textbf{0.83} & \textbf{0.92} & \textbf{0.84} & \textbf{0.88}\\
  \end{tabular}
  \caption{Comparison between the full CREDO system and CREDO system without some modules. SS here is Semantic Similarity and SA is Sentiment Analysis}
  \label{tab:table4} 
\end{table*}
\begin{table*}
\begin{tabular}{c|cccccc} 
 Dataset&Ans. -Ans.&Headlines&Plagiarism&Postediting&Ques.-Ques.&Mean\\ [0.5ex] 
 \hline
 Baseline STS & 0.41 & 0.54&  0.70&\textbf{0.83}&0.04&0.51\\ 
 CREDO Similarity &\textbf{0.54}& \textbf{0.71} & \textbf{0.73} &  0.82  & \textbf{0.61}&\textbf{0.68}\\
 Improvement (\%) & 31.7 & 31.5 &4.3&-1.2&1425& 33.3\\
\end{tabular}
\caption {Correlation Score in STS 2016 Task-1 of CREDO semantic similarity module}
\label{tab:table3}
\end{table*}
\begin{table*}[t]\centering
  \begin{tabular}{c|cccccccc}
   & & True & False & Macro- & & Fake & Fake & Fake \\
  & Overall  & Claims & Claims & averaged & & Claims & Claims & Claims \\ 
Classifier & Accuracy & Accuracy & Accuracy & Accuracy & AUC & Precision & Recall & F1-score \\\hline
Quadratic DA & 61.6 & 63.5 & 59.5 & 64.5 & 0.67 & 0.63 & 0.58 & 0.60\\
Naive Bayes& 65.8 & 67.1 & 64.3 & 66.7 & 0.68 & 0.71 & 0.65 & 0.69\\
Decision Trees & 66.3 & 67.4 & 64.5 & 65.4 & 0.69 & 0.70 & 0.67 & 0.68\\
AdaBoost & 70.6 & 72.3 & 69.4 & 71.1 & 0.74 & 0.75 & 0.72 & 0.73\\
Random Forest& 78.2 & 80.1 & 79.8 & 80.8 & 0.81 & 0.82 & 0.80 & 0.81\\
\textbf{RBF-SVM} & 82.8 & 85.7 & \textbf{81.5} & 83.6 & \textbf{0.87} & \textbf{0.94} & \textbf{0.85} & \textbf{0.89}\\
\textbf{MLP-NN} & \textbf{83.3} & \textbf{86.2} & 81.2 & \textbf{83.7} & 0.83 & 0.92 & 0.84 & 0.88\\
\end{tabular}
\caption{Comparison between different classifiers.}
\label{tab:table2} 
\end{table*}

\subsection{CREDO System}
\label{sec:credo system}
We train and test the CREDO system on the Snopes dataset. The experiment is conducted for binary credibility classification. The \textit{mostly true} and \textit{true} are considered positive labels and \textit{mostly false} and \textit{false} are considered negative labels. However, for our second experiment, we conduct multi-class credibility classification. For this, \textit{true},\textit{mostly true},\textit{mostly false} and \textit{false} are considered separate labels in a decreasing order.

Accuracy, Precision, Recall and F1-score are the evaluation metrics of this experiment. For training the weights, different classifiers were used - Support Vector Methods(SVM) with RBF kernel \cite{amari1999improving} and Neural Networks (Multi-Layer Perceptron) \cite{riedmiller2014multi} had the best results. Evaluation metrics for multi-class SVM use methods from \cite{godbole2004discriminative}.

K-fold cross validation (K=5) was used for training and testing on Snopes dataset. The evaluation metrics obtained are averaged across all the evaluation sets. The results of experiments with different classifiers are given in table \ref{tab:table1}.

\subsection{Dependency of CREDO on its modules} \label{sec:dependencyonmodules}
CREDO is a system based on 5 modules, but not all of them contribute equally. So, this experiment studies each modules' contribution to the overall system. For this, the effect on the performance of CREDO due to the exclusion of modules is analyzed in contrast to the original system.

Semantic similarity has to be tested independently to observe its effect on the overall performance of CREDO. The performance of semantic similarity module is tested on standard SemEval-STS 2016 English task.

\subsection{Dependency of CREDO on the classifier}\label{sec:credo classifier}
The overall problem is of classification and hence the choice of classifier has to be appropriate for the data points. Quadratic Discriminant Analysis (QDA) \cite{srivastava2007bayesian}, Gaussian Naive Bayes \cite{john1995estimating}, Decision Trees \cite{quinlan1986induction}, Random Forests (ensemble of Decision Trees) \cite{breiman2001random}, AdaBoost Classifier \cite{mathanker2011adaboost}, Support Vector Methods with Radial Basis Function kernel (SVM-RBF) \cite{amari1999improving} and Multi Layer Perceptron Neural Network (MLP-NN) \cite{riedmiller2014multi} were chosen for the experiments based on the variance of their application. All the evaluation metrics, tested for the original system, were tested for the classifiers as well.

\subsection{Evaluation of the Experiments}
\label{sec:evaluation of the experiments}
Table \ref{tab:table1} shows the comparative results of CREDO on Snopes dataset. CREDO outperforms the state-of-the-art approaches across all the metrics.
It is also observed that CREDO shows improvements over other approaches, even without the help of the Web of Trust module. Hence, the combination of Web of Trust and author credibility scoring modules helps in boosting the performance.

The results for the experiment on dependency of CREDO on modules (given in Section \ref{sec:dependencyonmodules}) is shown in table \ref{tab:table4}. The results show that ACS and WTS scores help the model but the improvements are insignificant compared to the enhancement in the metrics brought by the semantic similarity module. Including sentiment analysis module increments the accuracy, but it is minor compared to the improvement given by semantic similarity. At the same time, also not as insignificant as ACS and WTS scores. The results for the semantic similarity module are given in table \ref{tab:table3}. The results show the improvements the model had over the baseline system of the organizers in terms of Pearson Correlation score between the true similarity in the dataset and the scores of the given model.

The comparative results of the experiment with different classifiers (given in Section \ref{sec:credo classifier}) are given in table \ref{tab:table2}. As observed, the RBF-SVM model and MLP-NN model are better than the other architectures and have similar scores. Hence, these models were chosen for the main architecture.

\section{Conclusion}
\label{sec:conclusion}
We have proposed a neural network based approach for credibility analysis of unstructured text articles in an open-domain setting. We use a combination of relevant document retrieval techniques with semantic similarity, sentiment analysis and source reliability of articles then reporting the credibility score of the given input. Experiments on Snopes data demonstrate the effectiveness of our approach compared to the previous approaches to the problem.

Experiment on contribution of modules (given in Section \ref{sec:dependencyonmodules}) shows that the website and author reliability are helpful, but only to a limited extent. The accuracy scores had a difference of about 1\% . Semantic similarity and document retrieval are the major contributors to the system. The exclusion of the semantic similarity module resulted in a slash of 32.7\%. Sentiment Analysis contributed 7.1\% accuracy. Hence it can be concluded that the most important module is semantic similarity, followed by sentiment analysis, and then author, website scores.

The experiments also show that the choice of the classifier plays a major role in the output. Support Vector Methods with RBF kernel and Neural Network architecture (Multilayer Perceptron) give the best results for the problem. This is because the data points are not trivially classifiable and require non-linear classification. The neural network solves it by using multiple lines to classify the dataset whereas SVM utilizes the non-linear RBF kernel to address the problem.

As part of future work, we would like to apply this system to domains like social media. Also, we would like to enhance CREDO with more handcrafted features like the writing style.
\bibliographystyle{splncs03} 
\bibliography{main}

\end{document}